\documentclass[runningheads,a4paper]{llncs}
\usepackage{amssymb}
\usepackage{graphicx}
\usepackage{booktabs}
\usepackage{listings}
\usepackage{xcolor}
\usepackage{caption}
\usepackage{cite}
\usepackage{subcaption} 
\usepackage{color}
\usepackage{url}            

\definecolor{codegreen}{rgb}{0,0.6,0}
\definecolor{codegray}{rgb}{0.5,0.5,0.5}
\definecolor{codepurple}{rgb}{0.58,0,0.82}
\definecolor{backcolour}{rgb}{0.95,0.95,0.92}
\lstdefinestyle{mystyle}{
    backgroundcolor=\color{backcolour},   
    commentstyle=\color{codegreen},
    keywordstyle=\color{magenta},
    numberstyle=\tiny\color{codegray},
    stringstyle=\color{codepurple},
    basicstyle=\ttfamily\footnotesize,
    breakatwhitespace=false,         
    breaklines=true,                 
    captionpos=b,                    
    keepspaces=true,                 
    numbers=left,                    
    numbersep=5pt,                  
    showspaces=false,                
    showstringspaces=false,
    showtabs=false,                  
    tabsize=2
}
\usepackage[frozencache=true,cachedir=minted-cache]{minted}

\usepackage{enumitem}
\usepackage[colorlinks]{hyperref}

\urldef{\mailsa}\path|{g.anthony.reina@intel.com}|    
\newcommand{\keywords}[1]{\par\addvspace\baselineskip
\noindent\keywordname\enspace\ignorespaces#1}

\institute{
Intel Corporation, Santa Clara, CA 95052, USA
\and
Center for Biomedical Image Computing and Analytics, University of Pennsylvania, Philadelphia, PA, USA
\and
Department of Radiology, Perelman School of Medicine, University of Pennsylvania, Philadelphia, PA, USA
\and
Department of Pathology and Laboratory Medicine, Perelman School of Medicine, University of Pennsylvania, Philadelphia, PA, USA\\
\textsuperscript{*} Corresponding authors \email{\{prashant.shah@intel.com, sbakas@upenn.edu\}}
}

\begin{document}
\mainmatter

\title{OpenFL: An open-source framework for Federated Learning}

\titlerunning{OpenFL}

\author{
G Anthony Reina\inst{1}
\and
Alexey Gruzdev\inst{1}
\and
Patrick Foley\inst{1}
\and
Olga Perepelkina\inst{1}
\and
Mansi Sharma\inst{1}
\and
Igor Davidyuk\inst{1}
\and
Ilya Trushkin\inst{1}
\and
Maksim Radionov\inst{1}
\and
Aleksandr Mokrov\inst{1}
\and
Dmitry Agapov\inst{1}
\and
Jason Martin\inst{1}
\and
Brandon Edwards\inst{1}
\and
Micah J. Sheller\inst{1}
\and
Sarthak Pati\inst{2,3,4}
\and
Prakash Narayana Moorthy\inst{1}
\and
Shih-han Wang\inst{1}
\and 
Prashant Shah\inst{1,*}
\and
Spyridon Bakas\inst{2,3,4,*}
}

\authorrunning{G A Reina \textit{et al.}}

\maketitle

\begin{abstract}
Federated learning (FL) is a computational paradigm that enables organizations to collaborate on machine learning (ML) projects without sharing sensitive data, such as, patient records, financial data, or classified secrets. Open Federated Learning (OpenFL)\footnote{\url{https://github.com/intel/openfl}} is an open-source framework for training ML algorithms using the data-private collaborative learning paradigm of FL. OpenFL works with training pipelines built with both TensorFlow and PyTorch, and can be easily extended to other ML and deep learning frameworks. Here, we summarize the motivation and development characteristics of OpenFL, with the intention of facilitating its application to existing ML model training in a production environment. Finally, we describe the first use of the OpenFL framework to train consensus ML models in a consortium of international healthcare organizations, as well as how it facilitates the first computational competition on FL.

\end{abstract}

\section{Motivation}
\label{sec:motivation}

In the last decade, artificial intelligence (AI\footnote{\url{https://www.intel.com/content/www/us/en/artificial-intelligence/overview.html}}) has flourished due to greater access to data \cite{paullada2020data}. However, accessing vast and importantly diverse annotated datasets remains challenging because the underlying data are either too large or too sensitive to transmit to a centralized server for training a machine learning (ML) model \cite{sheller2020}.

\begin{figure}
  \centering
  \includegraphics[width=0.92\textwidth]{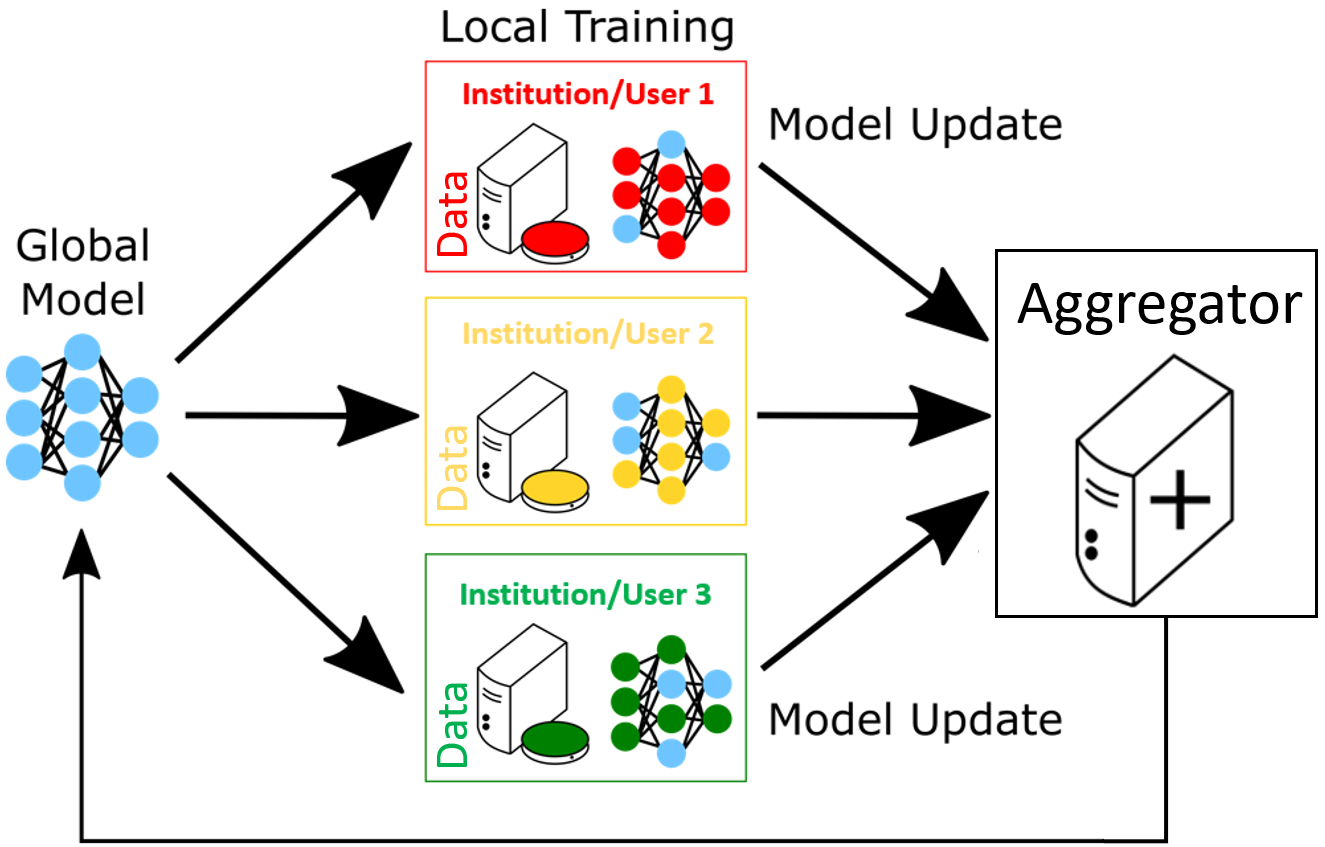}
  \caption{Federated Learning Workflow. A global model (left) is sent to different institutions/users for training on their local data. The model rather than the data is moved around the network. An aggregator node (right) combines model updates to generate a new global model that is sent back to the local institutions/users for further training. For details, confer with ~\cite{sheller2020,mcmahan2017communicationefficient}.}
  \label{fig:federated_learning}
\end{figure}

Federated learning (FL)\footnote{\url{https://en.wikipedia.org/wiki/Federated_learning}} is a collaborative computational paradigm that enables organizations to collaborate on data science projects without sharing sensitive information, such as patient records (protected health information), financial transactions, or protected secrets \cite{sheller2020,sheller2019,yang2019federated,mcmahan2017communicationefficient}. The basic premise behind FL, is that the AI model moves to meet the data, instead of the data moving to meet the model (that represents the current paradigm for multi-site collaborations) (Fig. \ref{fig:federated_learning}).

Recently, institutions have introduced FL deployments to train AI models for the healthcare\footnote{\url{https://intel.ly/3oaaE7Y}} and financial\footnote{\url{https://intel.ly/33FoJRm}} sectors. The goal is to allow greater access to larger and more diverse datasets without violating privacy laws, such as the Health Insurance Portability and Accountability Act (HIPAA)\footnote{\url{https://www.hhs.gov/hipaa/index.html}} of the United States \cite{annas2003hipaa} and the General Data Protection Regulation (GDPR) \footnote{\url{https://en.wikipedia.org/wiki/General_Data_Protection_Regulation}} \cite{voigt2017eu} of the European Union. Models trained using a FL approach can achieve similar levels of accuracy as models trained using a centralized learning approach \cite{sheller2019,sheller2020,mcmahan2017communicationefficient,suzumura2019federated}. 

In this paper, we outline our contribution of a new, open-source FL framework to the community. We overview the design and use of the framework and describe how to convert existing ML (and particularly deep learning) training instances into a federated training pipeline. Finally, we show how this new FL framework is being used to train a consensus ML model to detect and quantify boundaries of brain cancer, in a consortium of international healthcare organizations, as well as how it is used to facilitate the first computational competition on FL.

\section{The `Open Federated Learning' framework}

\subsection{Synopsis}

Open Federated Learning (OpenFL)\footnote{\url{https://github.com/intel/openfl}} is a software platform for federated learning (FL) that was initially developed as part of a collaborative research project between Intel Labs and the University of Pennsylvania on FL for healthcare, and continues to be developed for general-purpose real-world applications by Intel and the open-source community in GitHub\footnote{\url{https://github.com/intel/openfl}}. Although the initial use case was in healthcare, the OpenFL project is designed to be agnostic to the use-case, the industry, and the ML framework. The code is open-source, mostly in Python, and distributed via pip\footnote{\url{https://pypi.org/project/openfl/}}, conda, and Docker packages. The product allows developers to train ML models on the nodes of remote data owners (aka collaborators). The ML model is trained on the hardware at the collaborator node. Current examples are artificial neural networks trained using either TensorFlow \cite{tensorflow} or PyTorch \cite{pytorch}. Other ML model libraries and neural network training frameworks can be supported through an extensible mechanism. The data used to train the model remains at the collaborator node at all times; only the model weight updates and metrics are shared to the model owner via the aggregator node. A FL plan is used to describe the configuration and workflow. This FL plan is shared among all nodes in the federation to define the rules of the federation. Thanks to \cite{bonawitz2019federated} for the term \textit{FL plan}, though as OpenFL has been designed for a different trust model (multi-institutional), the OpenFL plan is agreed upon by all parties before the workload begins, as opposed to the design in \cite{bonawitz2019federated} which delivers the FL plan at runtime (as befits that system's design goals).

\begin{figure}
  \centering
  \includegraphics[width=0.8\textwidth]{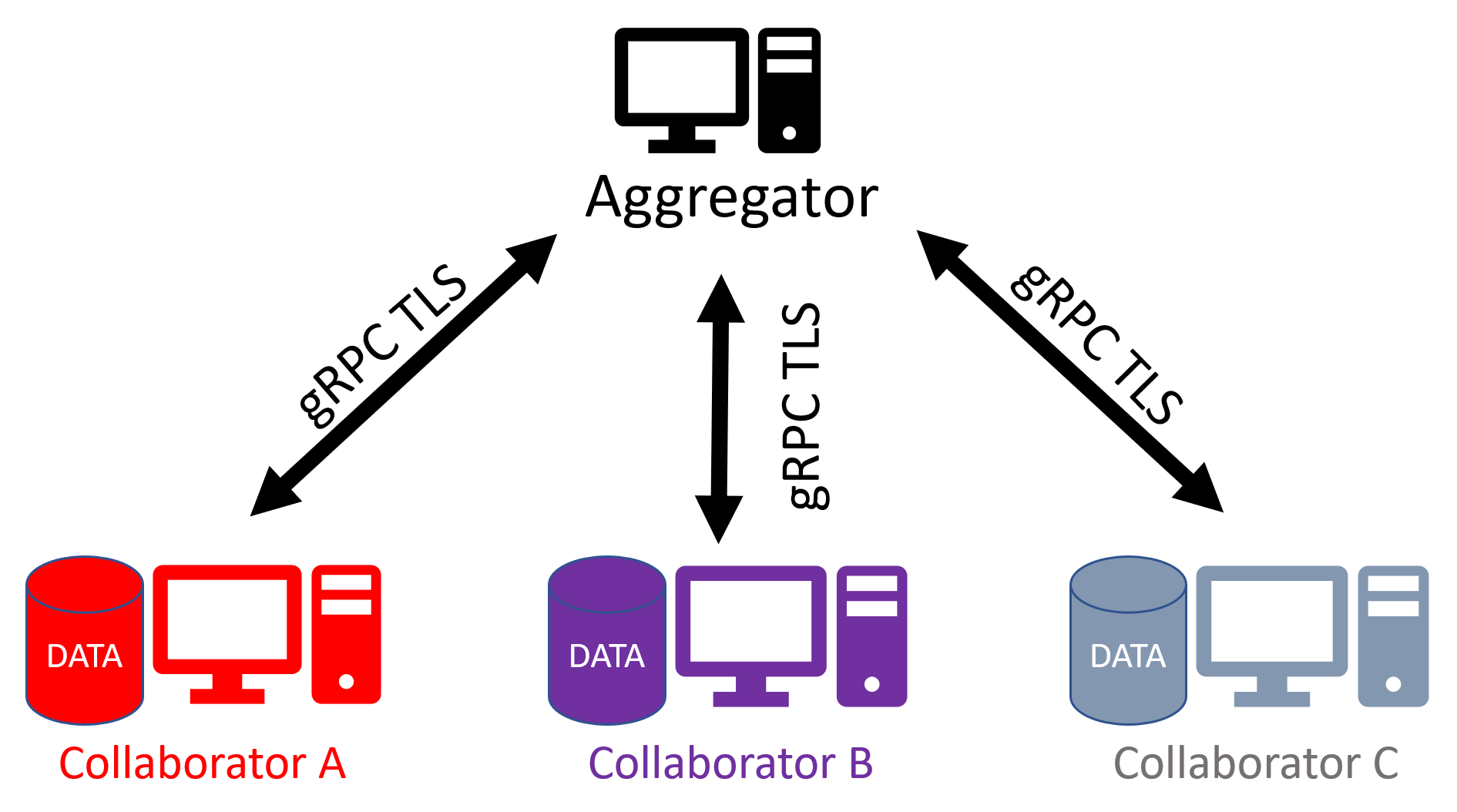}
  \caption{The OpenFL network topology. The federation is a star topology with two types of nodes: collaborators and aggregators. The data of a collaborator remains within that node for local training. The dataset never leaves the collaborator node. Instead, model updates from each collaborator node are sent to an aggregator node so that they can be combined into a global consensus model. The global model is returned to the collaborator nodes for a further round of local training. Collaborators connect with the aggregator through remote procedure calls over mutually-authenticated TLS connections.}
  \label{fig:openfl_simple_architecture}
\end{figure}

\begin{figure}
  \centering
  \includegraphics[width=0.9\textwidth]{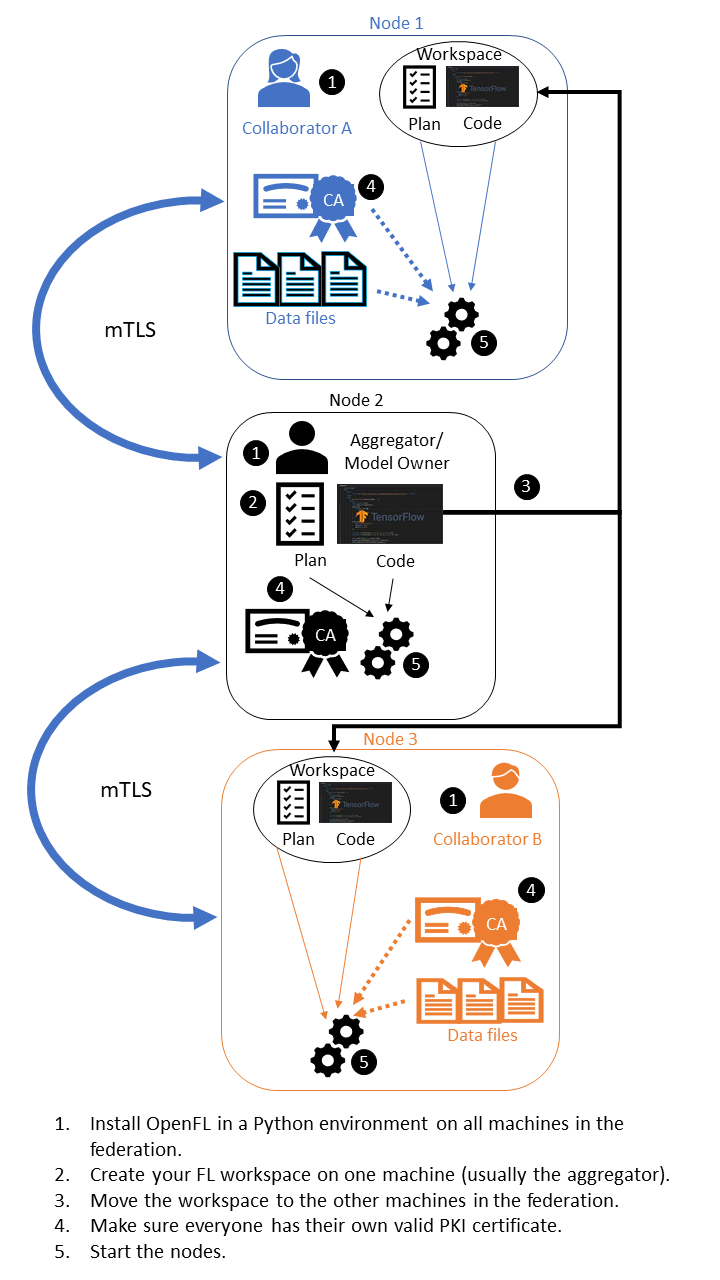}
  \caption{A high-level overview of Open Federated Learning (OpenFL). Note that once OpenFL is installed on all nodes of the federation and every member of the federation has a valid PKI certificate, all that is needed to run an instance of a federated workload is to distribute the workspace to all federation members and then run the command to start the node.}
  \label{fig:openfl_workflow}
\end{figure}

The high-level workflow is shown in the Fig. \ref{fig:openfl_workflow}. In this example, we are distributing a TensorFlow model, but PyTorch and other frameworks are handled in the same way. Note that once OpenFL is installed on all nodes of the federation and every member of the federation has a valid PKI certificate, all that is needed to run an instance of a federated workload is to distribute the workspace to all federation members and then run the command to start the node (\textit{e.g.} \verb|fx aggregator start|/\verb|fx collaborator start|). In other words, most of the work is setting up an initial environment (steps 1-4) on all of the federation nodes that can be used across new instantiations of federations.

\subsection{Architecture}

Fig. \ref{fig:openfl_simple_architecture} shows the architecture for the OpenFL network topology. Each participant in the federation is defined as either a \textit{collaborator} or an \textit{aggregator} node. A collaborator node contains the dataset that is owned by that participant. The hardware of that collaborator node is used to train the ML model locally. The dataset never leaves the collaborator node. An aggregator node is a compute node that is trusted by each collaborator node. Collaborator nodes connect directly to the aggregator node in a star topology. The aggregator connects to the collaborator nodes through remote procedure calls (gRPC\footnote{\url{https://grpc.io/}} \cite{grpc}) via a mutually-authenticated transport layer security (TLS)\footnote{\url{https://en.wikipedia.org/wiki/Transport_Layer_Security}} network connection. Sensitive data such as tasks, model and optimizer weights, and aggregated metrics pass between the collaborator and the aggregator nodes over this encrypted channel.

\begin{figure}
  \centering
  \includegraphics[width=\textwidth]{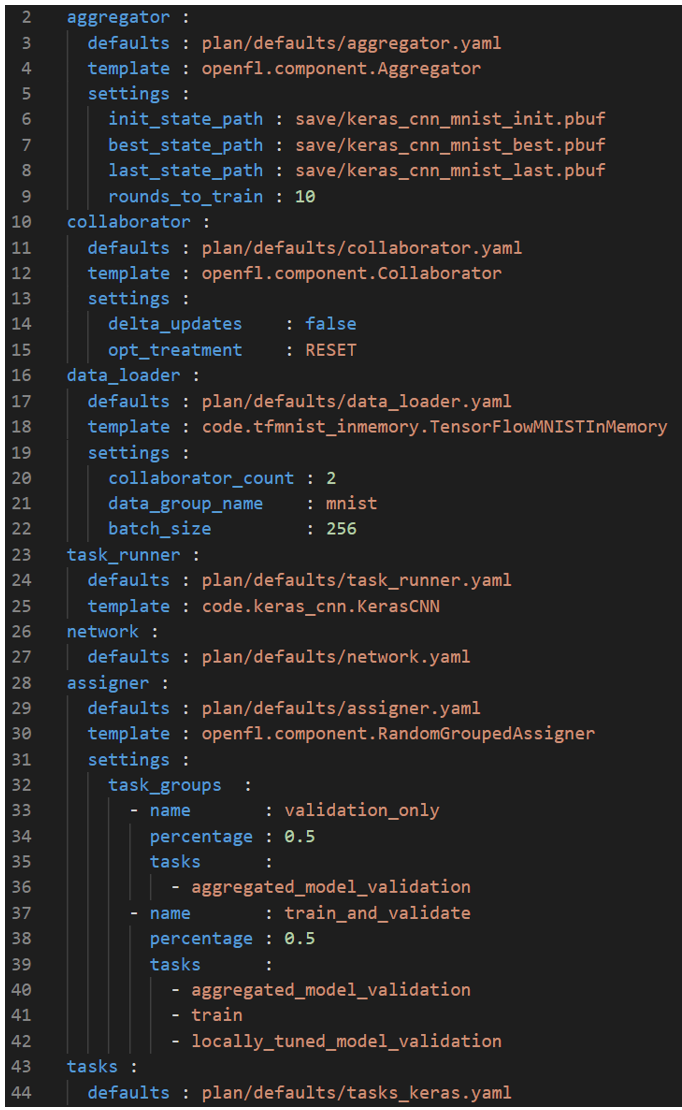}
  \caption{A Federated learning (FL) plan is a YAML file that defines the tasks and parameters required to coordinate and execute a federation. The FL plan is shared to all participants within the federation. It defines the collaborator and aggregator settings as well as the remote procedure calls for this federation. Note that in this example lines 28-42 of the FL plan assign different tasks ("validation\_only" and "train\_and\_validate") to random collaborators during a given round of the federation.}
  \label{fig:fl_plan}
\end{figure}

The coordination and execution of a given federation is defined by the FL plan (Fig. \ref{fig:fl_plan}). The FL plan is a YAML\footnote{\url{https://yaml.org/}} file that is shared with all the participants of a given federation. It defines the collaborator and aggregator settings, such as batch size, IP address, and rounds to train. It also specifies the remote procedure calls for the given federation tasks.

\begin{figure}
  \centering
  \includegraphics[width=\textwidth]{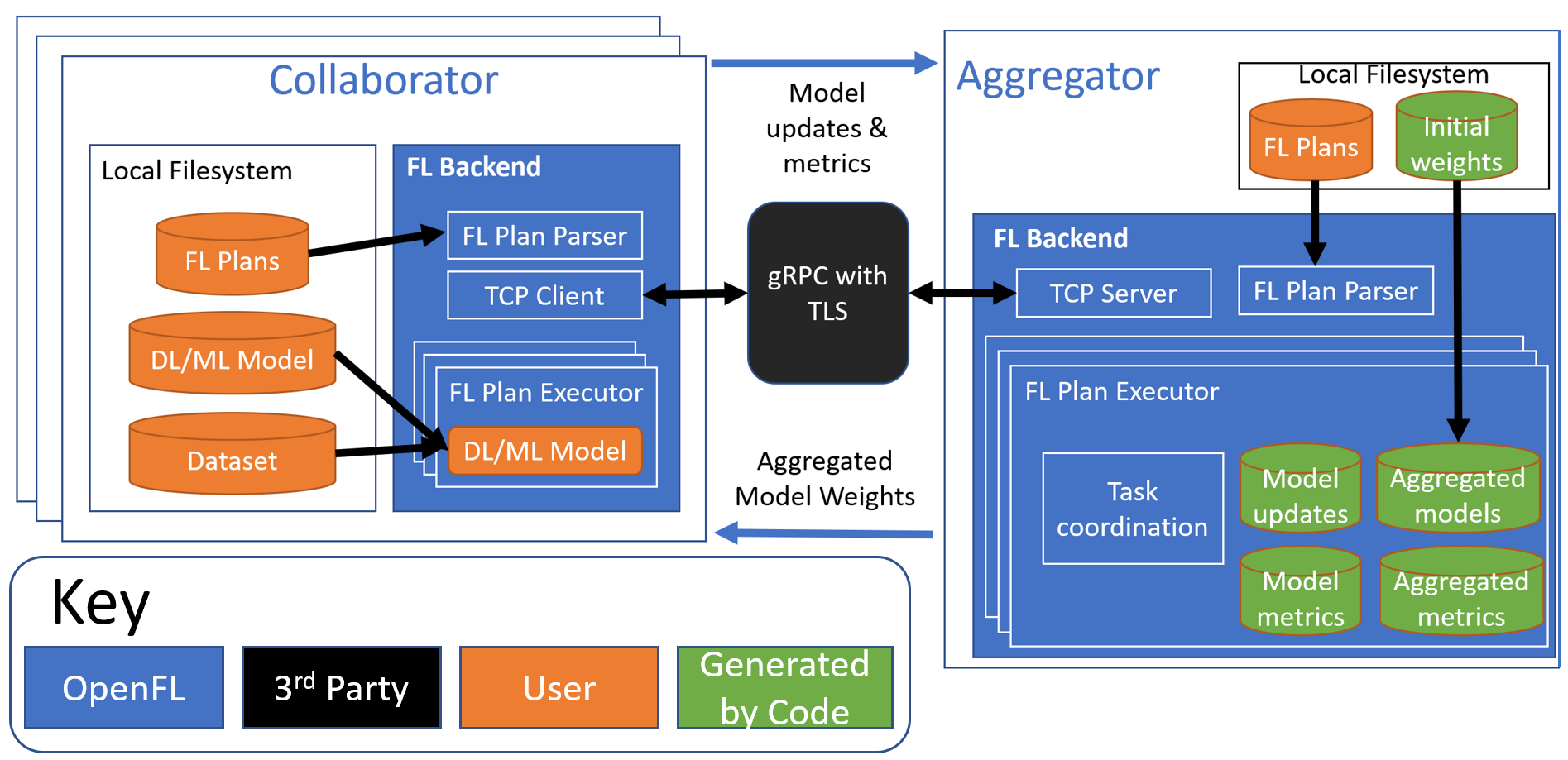}
  \caption{The OpenFL software components. The collaborator contains the federation plan (FL Plan), ML model, and local dataset. These components are created by the developer (orange). The OpenFL backend (blue) connects the collaborator with the aggregator node via a mutually-authenticated TLS connection. The OpenFL backend (blue) on the aggregator sends remote procedure calls to the collaborator and receives model and metric updates (green) for aggregation.}
  \label{fig:openfl_architecture}
\end{figure}

Fig. \ref{fig:openfl_architecture} shows the software components of the OpenFL framework. The collaborator contains the federation plan (FL Plan), the ML model code, and the local dataset. The FL plan and model code are manually shared with each participant prior to the start of the federation using an export command in the OpenFL command-line interface (CLI) as described later. The OpenFL backend allows the aggregator node to send remote procedure calls to the collaborators which instructs them to execute tasks as defined in the FL plan, such as ML model training and validation. When they have completed their tasks, then the collaborators report the updated model weights (and aggregated metrics, such as model accuracy and local dataset size) to the aggregator. The aggregator combines the updates from the collaborators into a global consensus model as described by the algorithm specified in the FL plan. The aggregator then sends the weights of the new global model back to the collaborators for an additional round of tasks (Fig. \ref{fig:federated_learning}). This process continues until all rounds have been completed as specified in the FL plan.

\subsection{Security}\label{Security}
 FL addresses issues of the current paradigm for multi-institutional collaborations based on data pooling, but also introduces new privacy, security, and confidentiality challenges \cite{kairouz2021advances}. AI model builders may wish to protect their model intellectual properties (IP) as the model trains in decentralized environments, while data holders/contributors would like to ensure that their data cannot be extracted by inspecting the model weights over federated rounds. Initially developed in the Intel Labs Security and Privacy Research lab, the OpenFL design prioritizes key security concepts such as narrow interfaces, code reuse, open-source code, simplified information security reviews, and code design fit for running on trusted compute hardware, such as a trusted execution environment (TEE).

\subsubsection{PKI Certificates}
OpenFL uses mutually authenticated\footnote{\url{https://en.wikipedia.org/wiki/Mutual_authentication}} transport layer security (TLS)\footnote{\url{https://en.wikipedia.org/wiki/Transport_Layer_Security}} connections. To establish the connection, a valid public key infrastructure certificate\footnote{\url{https://en.wikipedia.org/wiki/Public_key_infrastructure}} signed by a trusted certificate authority (CA) must be provided by all participants. OpenFL provides a method for creating a certificate authority and generating X.509\footnote{\url{https://en.wikipedia.org/wiki/X.509}} certificates, but this mechanism is only intended for \textbf{non-production testing}. In production settings, it is recommended that a trusted certificate authority generates the PKI certificates. The minimum recommended certificates are \textit{RSA SHA-384 3072-bit} or \textit{ECDSA secp384r}. 

\subsubsection{Trusted Execution Environments}
Trusted Execution Environments (TEEs) provide hardware mechanisms to execute code with various security properties. For FL, the three key security properties we want from a TEE are 1) \textit{confidentiality} of the execution to mitigate attacks such as copying model IP out of memory as the training process executes, 2) \textit{integrity} of the execution to mitigate attacks that alter the behavior of the code, and 3) \textit{remote attestation} of the execution, wherein a TEE can provide some measure of proof regarding the initial execution state to a remote party to ensure that the remote party is interacting with the intended code on the intended hardware \cite{kairouz2021advances}. The latest Intel Scalable Xeon processors provide a TEE on the host CPU via the Intel Software Guard Extensions (Intel SGX) that can provide these three security properties at near-native speed, supporting memory sizes necessary for large deep learning models. OpenFL originates from the same research lab that created Intel SGX, and has been designed to properly leverage TEEs from the outset. Instructions for how to run OpenFL with SGX are outside the scope of this paper.

Intel SGX applications run in user-mode (as opposed to kernel-mode), thus in order to access kernel functions within the TEE, we have used the open-source library-OS Graphene with SGX (GSGX)\cite{gsgx}. GSGX allows us to run OpenFL code with Intel SGX without any modification to the OpenFL code, and provides mechanisms such as file system encryption and integrity.


\subsection{Installation}

\subsubsection{Baremetal}

\begin{figure}
  \centering
  \includegraphics[width=\textwidth]{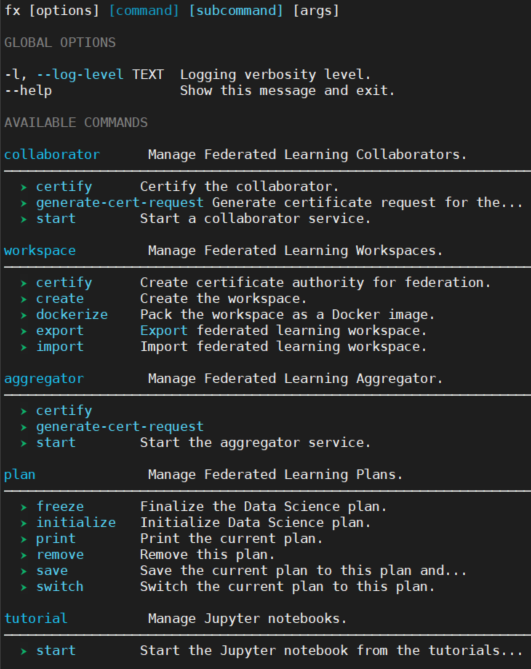}
  \caption{OpenFL's fx Command Line Interface}
  \label{fig:openfl_cli}
\end{figure}

OpenFL has been validated on Ubuntu Linux\footnote{\url{https://ubuntu.com/}} 16.04 and 18.04. To install the latest version, it is recommended to create a Python virtual environment using either ``venv''\footnote{\url{https://docs.python.org/3/tutorial/venv.html}} or ``conda''\footnote{\url{https://www.anaconda.com/products/individual}}. Within the virtual environment, run the following command to install directly from ``PyPI''\footnote{\url{https://pypi.org/project/openfl/}}:

\begin{minted}[xleftmargin=\parindent]{bash}
pip install openfl
\end{minted}

Alternatively, a Python wheel can be generated by downloading or cloning the source code from \url{https://github.com/intel/openfl} and running the commands:

\begin{minted}[xleftmargin=\parindent,linenos]{bash}
./scripts/build_wheel.sh
pip install ./dist/openfl-*.whl
\end{minted}

Once OpenFL has been successfully installed, a new CLI called \verb|fx| is available to develop federations. The \verb|fx| command help page is listed in Fig. \ref{fig:openfl_cli}.

\subsubsection{Docker}

If Docker is installed and running, then simply run the command:

\begin{minted}[xleftmargin=\parindent]{bash}
docker pull intel/openfl
\end{minted}

If you would prefer to build an image from a specific commit or branch, run the following commands:

\begin{minted}[xleftmargin=\parindent,linenos]{bash}
git clone https://github.com/intel/openfl.git
cd openfl
./scripts/build_base_docker_image.sh
\end{minted}

To run the base Docker container:

\begin{minted}[xleftmargin=\parindent]{bash}
docker run -it --network host openfl
\end{minted}

\subsection{Running a Federation}

OpenFL has two methods for developing federations: the Python API and the \verb|fx| CLI. The CLI is considered the better path for scaling federations within a production environment. The Python API is easier to understand for the data scientist who is working with OpenFL for the first time. Nevertheless, the OpenFL tutorials and demos \footnote{\url{https://openfl.readthedocs.io/en/latest/running_the_federation.notebook.html}} allow users to quickly learn both methods.

\subsubsection{Python API}
Once OpenFL is installed in a Python virtual environment, type the command:

\begin{minted}[xleftmargin=\parindent]{bash}
fx tutorial start
\end{minted}

which will start a Jupyter Notebook server\footnote{\url{https://jupyter.org}} that hosts the OpenFL Python API tutorials. Open a web browser and start the notebook tutorial \textbf{Federated\_Keras\_MNIST\_Tutorial.ipynb}. 

\begin{listing}[ht]
\begin{minted}[
frame=lines,
fontsize=\footnotesize,
linenos
]{python}
from tensorflow.keras import Sequential
from tensorflow.keras.layers import Dense

import openfl.native as fx
from openfl.federated import FederatedModel, FederatedDataSet

# Setup default workspace, logging, etc.
fx.init('keras_cnn_mnist')

"""
 Define the data loader and pre-processing steps here
"""

fl_data = FederatedDataSet(train_images, train_labels,
          valid_images, valid_labels,
          batch_size=32,
          num_classes=classes)

def build_model(feature_shape,classes):
    # Defines the MNIST model
    model = Sequential()
    model.add(Dense(64, input_shape=feature_shape, activation='relu'))
    model.add(Dense(64, activation='relu'))
    model.add(Dense(classes, activation='softmax'))

    model.compile(optimizer='adam',
                  loss='categorical_crossentropy',
                  metrics=['accuracy'],)
    return model

fl_model = FederatedModel(build_model, data_loader=fl_data)

collaborator_models = fl_model.setup(num_collaborators=2)
collaborators = {'one':collaborator_models[0], 
                 'two':collaborator_models[1]}

final_fl_model = fx.run_experiment(collaborators,
                  override_config={'aggregator.settings.rounds_to_train':5})
\end{minted}

\caption{Python API example}
\label{listing:1}
\end{listing}

\begin{figure}
  \centering
  \includegraphics[angle=90,origin=c,width=0.92\textwidth]{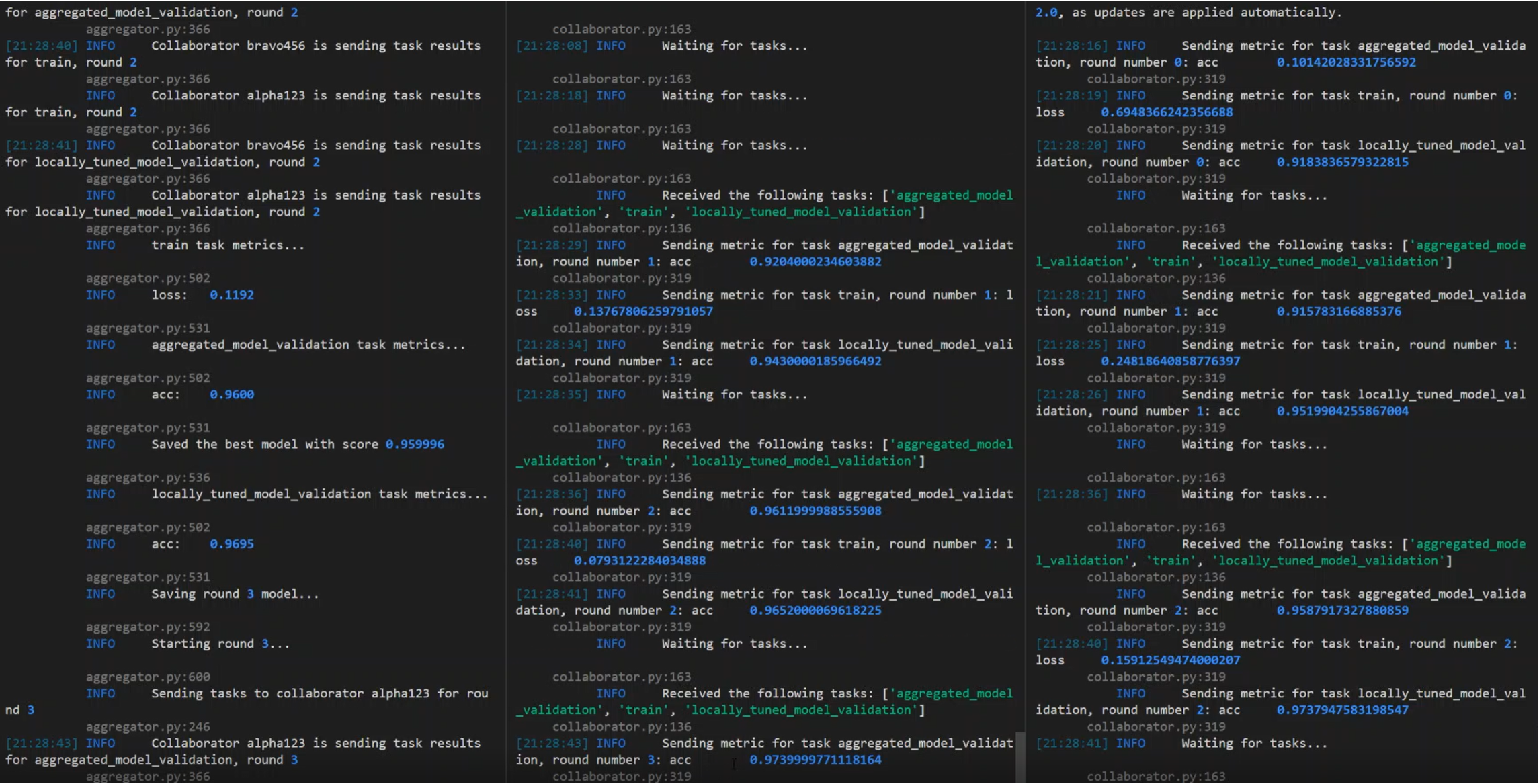}
  \caption{The OpenFL logs from the aggregator (left) and two collaborators (middle, right) during the Keras MNIST tutorial.}
  \label{fig:openfl_logs}
\end{figure}

Listing \ref{listing:1} shows a simplified version of the tutorial. Lines 4-5 import the OpenFL \verb|fx| function and the \textit{FederatedDataSet} and \textit{FederatedModel} classes. In this example, the \textit{FederatedDataSet} class will be used to wrap the MNIST \cite{mnist} dataset and shard it equally across the collaborators. Obviously, in a real-world case, the dataset would already exist on the individual collaborators and would need no sharding. On line 31, the \textit{FederatedModel} class is used to instantiate an FL model. Note the \verb|fx| commands on lines 8 and 37. Here the \verb|fx| command is used to initialize an FL workspace directory for the code and then run the FL experiment. An example of the logs for the experiment is shown in Fig. \ref{fig:openfl_logs}.

\subsubsection{Experimental: Interactive Python API}

An alternate OpenFL Python API Layer is being developed to further simplify defining FL experiments. It is anticipated that this approach will be the preferred API for developers in the future to create multi-node FL workloads. The interactive API allows creating and launching experiments from a single entry point - a Jupyter notebook or a Python script. Users of the Interactive API can operate on Python objects and use them in FL experiments. Defining an experiment includes setting up several interface entities and experiment parameters.

Tutorial notebooks with the Interactive API use cases are available as a part of OpenFL repository and Listing \ref{listing:2} shows a simplified example. Lines 3-7 describe the Federation object setup; Federation API is the place to tune network parameters and settings that are specific for different collaborator nodes. On lines 10-45 interactive API components are used to register an ML model (with its optimizer), an FL task, and an FL data loader. Lines 48-56 are devoted to the Experiment API that uses the Python objects previously defined to compile all the parts and settings required to start a federation. This API can pack these components into a distributable archive and can start the aggregator to conduct the FL experiment.

\begin{listing}[ht]
\begin{minted}[
frame=lines,
fontsize=\footnotesize,
linenos
]{python}
# First, create a federation objects
from openfl.interface.interactive_api.federation import Federation
federation = Federation(central_node_fqdn: str, disable_tls: bool,
                        cert_chain: str, agg_certificate: str, agg_private_key: str)
federation.register_collaborators(col_data_paths=
    {"collaborator 1 name": "local data path",
     "collaborator 2 name": "local data path"})

# Register a model, tasks, and a dataloader
from openfl.interface.interactive_api.experiment import 
    TaskInterface, DataInterface, ModelInterface
# Model
MI = ModelInterface(model, optimizer, framework_plugin)

# Tasks
TI = TaskInterface()

task_settings = {
    'batch_size': 32,
    'some_arg': 228,
}
@TI.add_kwargs(**task_settings)
@TI.register_fl_task(model='my_model', data_loader='train_loader',
        device='device', optimizer='my_Adam_opt')
def foo_training_task(my_model, train_loader, my_Adam_opt, device, batch_size, some_arg=356)
    # Task body
    pass

# Dataloader
class FedDataset(DataInterface):
    def _delayed_init(self, data_path):
        # Do dataset preparations
        # data_path values were registered on the federation level
        

    def get_train_loader(self, **kwargs):
        # This method will be called before training tasks execution.
        # This method must return anything user expects to receive
        # in the training task with data_loader contract argument.
        pass

    def get_train_data_size(self):
        # return number of samples in local train dataset.
        pass
        
fed_dataset = FedDataset()

# Create an Experiment object
from openfl.interface.interactive_api.experiment import FLExperiment
fl_experiment = FLExperiment(federation=federation)

# FL experiment can automatically prepare the FL plan and the workspace archive
fl_experiment.prepare_workspace_distribution(model_provider=MI, task_keeper=TI,
                                             data_loader=fed_dataset, rounds_to_train=5, \
                                             opt_treatment='CONTINUE_GLOBAL')
# Start an aggregator with initial model weights
fl_experiment.start_experiment(model_provider=MI)

# Move the workspace archive to collaborator nodes and run collaborator processes
\end{minted}

\caption{Experimental Interactive Python API example}
\label{listing:2}
\end{listing}

\subsubsection{Command Line Interface (fx CLI)}

Federation can also be run using individual \verb|fx| CLI commands. Here are a series of steps to run a federation experiment with \verb|fx| CLI:\newline \newline 

1. Make sure that you have initialized the virtual environment.\newline

2. Create a workspace for the new federation project 
\begin{minted}[xleftmargin=\parindent]{bash}
fx workspace create --prefix ${WORKSPACE_PATH} 
    --template ${WORKSPACE_TEMPLATE}
\end{minted}
where \textit{--prefix} is the directory to create the workspace and \textit{--template} is the template that can be chosen from a list of pre-existing templates that can be found by running this command:

\begin{minted}[xleftmargin=\parindent]{bash}
fx workspace create --prefix ${WORKSPACE_PATH}
\end{minted}

3. Change to the workspace directory
\begin{minted}[xleftmargin=\parindent]{bash}
cd ${WORKSPACE_PATH} 
\end{minted}

4. Install workspace requirements
\begin{minted}[xleftmargin=\parindent]{bash}
pip install -r requirements.txt
\end{minted}

5. Initialize the plan and autopopulate the fully qualified domain name (FQDN) of the aggregator node.
\begin{minted}[xleftmargin=\parindent]{bash}
fx plan initialize
\end{minted}
Although it is possible to train models from scratch, it is assumed that in 
many cases the federation may perform fine-tuning of a previously-trained model. 
For this reason, the pre-trained weights for the model will be stored within protobuf 
files on the aggregator and passed to the collaborators during initialization. 
As seen in the YAML file, the protobuf file with the initial weights is expected 
to be found in the file \textit{\$\{WORKSPACE\_TEMPLATE\}\_init.pbuf}.
For this example, however, by running the command above, we’ll just create an initial 
set of random model weights that are put into that file.
The FQDN is embedded within the plan so the collaborators know the externally accessible aggregator server address to connect to. If you face connection issues with the autopopulated FQDN in the plan, this value can be overridden with the \textit{-a} flag, for example:
\begin{minted}[xleftmargin=\parindent]{bash}
fx plan initialize -a aggregator-hostname.internal-domain.com
\end{minted}

\textbf{On the Aggregator node}\newline
Ensure that Python virtual environment is activated and OpenFL package is installed.\newline

6. Change directory to the path for your project’s workspace:
\begin{minted}[xleftmargin=\parindent]{bash}
cd ${WORKSPACE_PATH} 
\end{minted}

7. Run the Certificate Authority command. This will setup the Aggregator node as the Certificate Authority for the Federation. All certificates will be signed by the aggregator. Follow the command-line instructions and enter in the information as prompted. The command will create a simple database file to keep track of all issued certificates.
\begin{minted}[xleftmargin=\parindent]{bash}
fx workspace certify
\end{minted}

8. Run the aggregator certificate creation command, replacing AFQDN with the actual fully qualified domain name (FQDN) for the aggregator machine.
\begin{minted}[xleftmargin=\parindent]{bash}
fx aggregator generate-cert-request --fqdn AFQDN
\end{minted}
If you omit the \textit{--fdqn} parameter, then \verb|fx| will automatically use the FQDN of the current node assuming the node has been correctly set with a static address.\newline

9. Run the aggregator certificate signing command, replacing AFQDN with the actual fully qualified domain name (FQDN) for the aggregator machine.
\begin{minted}[xleftmargin=\parindent]{bash}
fx aggregator certify --fqdn AFQDN
\end{minted}
This node now has a signed security certificate as the aggregator for this new federation. You should have the given files:\\
certificate chain (\textit{WORKSPACE.PATH/cert/cert\_chain.crt})\\
aggregator certificate (\textit{WORKSPACE.PATH/cert/server/agg\_AFQDN.crt})\\
aggregator key (\textit{WORKSPACE.PATH/cert/server/agg\_AFQDN.key})\newline

\textbf{Exporting the workspace}\newline

10. Export the workspace so that it can be imported to the collaborator nodes.
\begin{minted}[xleftmargin=\parindent]{bash}
fx workspace export
\end{minted}
The export command will archive the current workspace (as a zip) and create a \textit{requirements.txt} file of the current Python packages in the virtual environment. Transfer this zip file to each collaborator node.\newline

\textbf{On the Collaborator nodes}\newline

Before you run the federation make sure you have activated a Python virtual environment and installed the OpenFL package.\newline

11. Make sure you have copied the workspace archive (.zip) from the aggregator node to the collaborator node.\newline

12. Import the workspace archive
\begin{minted}[xleftmargin=\parindent]{bash}
fx workspace import --archive WORKSPACE.zip
\end{minted}
where \textit{WORKSPACE.zip} is the name of the workspace archive. This will unzip the workspace to the current directory and install the required Python packages within the current virtual environment.\newline

13. For each test machine you want to run collaborators on, we create a collaborator certificate request to be signed by the certificate authority, replacing \textit{COL.LABEL} with the label you’ve assigned to this collaborator. Note that this does not have to be the FQDN. It can be any unique alphanumeric label.
\begin{minted}[xleftmargin=\parindent]{bash}
fx collaborator generate-cert-request -n COL.LABEL
\end{minted}

The creation script will also ask you to specify the path to the data. For example, if using the MNIST dataset (from keras\_cnn\_mnist template), simply enter the an integer that represents which shard of MNIST to use on this Collaborator. For the first collaborator enter 1. For the second collaborator enter 2. This will create these three files:\\
Collaborator CSR (\textit{WORKSPACE.PATH/cert/client/col\_COL.LABEL.csr})\\ Collaborator key (\textit{WORKSPACE.PATH/cert/client/col\_COL.LABEL.key})\\ Collaborator CSR Package (\textit{WORKSPACE.PATH/col\_COL.LABEL\_to\_agg\_cert\\
\_request.zip})\\
Only the Collaborator CSR Package file needs to be sent to the certificate authority to be signed (for example, aggregator in this case).\newline

14. On the Aggregator node (i.e. the Certificate Authority for this demo), run the following command:
\begin{minted}[xleftmargin=\parindent,breaklines=True]{bash}
fx collaborator certify --request-pkg    /PATH/TO/col_COL.LABEL_to_agg_cert_request.zip
\end{minted}
where \textit{/PATH/TO/col\_COL.LABEL\_to\_agg\_cert\_request.zip} is the path to the package containing the \textit{.csr} file from the collaborator. The Certificate Authority will sign this certificate for use in the Federation.\newline

15. The previous command will package the signed collaborator certificate for transport back to the Collaborator node along with the \textit{cert\_chain.crt} needed to verify certificate signatures. The only file needed to send back to the Collaborator node is the following:\newline
\textit{WORKSPACE.PATH/agg\_to\_col\_COL.LABEL\_signed\_cert.zip}\newline

16. Back on the Collaborator node, import the signed certificate and certificate chain into your workspace:
\begin{minted}[xleftmargin=\parindent,breaklines=True]{bash}
fx collaborator certify --import /PATH/TO/agg_to_col_COL.LABEL_signed_cert.zip
\end{minted}

\textbf{Starting the federation}\newline
17. On the aggregator node, start the federation:
\begin{minted}[xleftmargin=\parindent]{bash}
fx aggregator start
\end{minted}
At this point, the aggregator is running and waiting for the collaborators to connect. When all of the collaborators connect, the aggregator starts training. When the last round of training is complete, the aggregator stores the final weights in the protobuf file that was specified in the YAML file (in this case \textit{save/\${WORKSPACE\_TEMPLATE}\_latest.pbuf}).\newline

18. On each of the collaborator nodes, start the collaborator:
\begin{minted}[xleftmargin=\parindent]{bash}
fx collaborator start -n COLLABORATOR.LABEL
\end{minted}
where \textit{COLLABORATOR\_LABEL} is the label for this collaborator.\newline

19. Repeat this for each collaborator in the federation. Once all collaborators have joined, the aggregator will start and you will see log messages describing the progress of the federated training.

\section{Use Cases}

    \subsection{Federated Tumor Segmentation Initiative}
    \label{sec:fets}
    
\begin{figure}
      \centering
      \includegraphics[width=\textwidth]{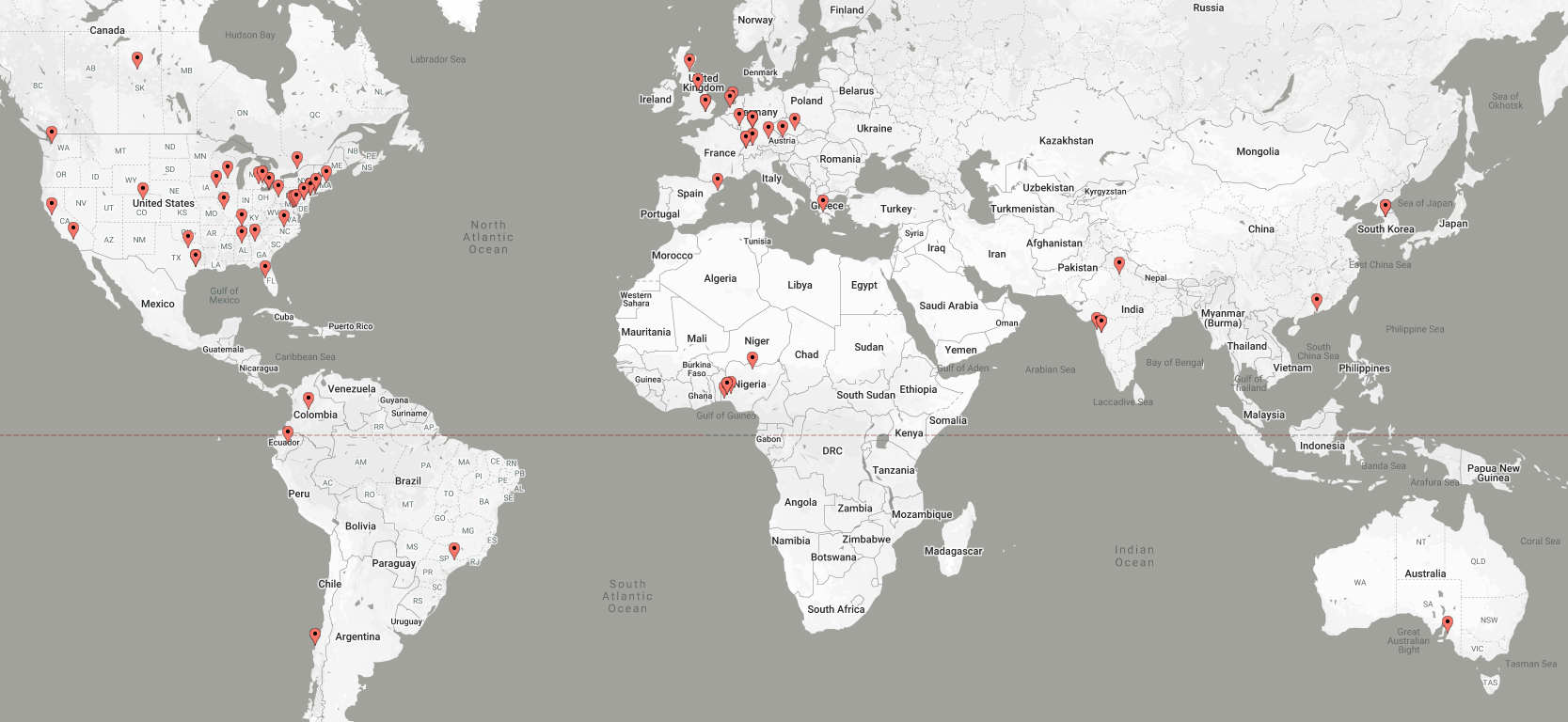}
      \caption{The collaborative network of the first FeTS federation.}
      \label{fig:fets_map}
    \end{figure}

    The Federated Tumor Segmentation (FeTS) initiative describes an on-going development of the largest international federation of healthcare institutions aiming at gaining knowledge for tumor boundary detection from ample and diverse patient populations without sharing any patient data. To facilitate this initiative, a dedicated open-source platform with a user-friendly graphical user interface was developed aiming at: i) bringing state of the art pre-trained segmentation models of numerous algorithms \cite{pati2021gandlf} and label fusion approaches \cite{labelfusion}, closer to clinical experts and researchers, thereby enabling easy quantification of new radiologic scans and comparative evaluation of new algorithms, and ii) allowing multi-institutional collaborations via FL by leveraging OpenFL to improve these pre-trained models without sharing patient data, thereby overcoming legal, privacy, and data-ownership challenges. FeTS has been initially deployed towards the task of brain tumor sub-region segmentation by partnering with $n=56$ clinical sites spread all around the world (Fig. \ref{fig:fets_map}).
    
    \subsection{First Computational Competition on Federated Learning}
    
    International challenges have become the \textit{de facto} standard for validating medical image analysis methods. However, the actual performance of even the winning algorithms on ``real-world'' clinical data often remains unclear, as the data included in these challenges are usually acquired in very controlled settings at few institutions. The seemingly obvious solution of just collecting increasingly more data from more institutions in such challenges does not scale well due to privacy and ownership hurdles (Section \ref{sec:motivation}).

    As the first challenge ever proposed for federated learning, the FeTS challenge 2021\footnote{\url{https://www.med.upenn.edu/cbica/fets/miccai2021/}} intends to address these hurdles towards both the creation and the evaluation of tumor segmentation models. Specifically, the FeTS 2021 challenge uses clinically acquired, multi-institutional MRI scans from the BraTS 2020 challenge \cite{brats_1,brats_2,brats_3}, as well as from various remote independent institutions included in the collaborative network of a real-world federation (Section \ref{sec:fets}). The challenge focuses on the construction and evaluation of a consensus model for the segmentation of intrinsically heterogeneous (in appearance, shape, and histology) brain tumors, namely gliomas. Compared to the BraTS challenge, the ultimate goal of the FeTS challenge is divided into the following two tasks:
    
    \begin{enumerate}
        \item \textbf{Task 1} (``Federated Training'') aims at effective weight aggregation methods for the creation of a consensus model given a pre-defined segmentation algorithm for training, while also (optionally) accounting for network outages.
        \item \textbf{Task 2} (``Federated Evaluation'') aims at robust segmentation algorithms, given a pre-defined weight aggregation method, evaluated during the testing phase on unseen datasets from various remote independent institutions of the collaborative network of the \textbf{fets.ai} federation.
    \end{enumerate}

\section{Discussion}

Kaushal \textit{et al.} \cite{langlotz_bias} recommend that researchers need greater access to large and diverse datasets, in order to generate accurate models \cite{langlotz_bias}. Without this greater access, they argued, AI models may also have inherent biases and perpetual inequalities. For example, Larrazabal \textit{et al.} demonstrated that introducing a gender imbalance while training convolutional neural network model to detect disease from chest X-rays resulted in poor performance on the underrepresented gender \cite{gender_imbalance_ai}. This potential for bias is not limited to the healthcare sector. Buolamwini \textit{et al.} \cite{pmlr-v81-buolamwini18a} demonstrated that a lack of diversity in training data can lead to significant racial bias in facial detection algorithms. Coston \textit{et al.} described the harmful effects as a covariate shift in risk models for the financial sector \cite{coston_covariate}.

Federated learning (FL) is an attractive approach to training AI on large, diverse datasets requiring data privacy \cite{rieke2020,suzumura2019federated}. Although there is no inherent guarantee that accessing more data translates to accessing better data, it is certainly a step in the right direction toward improving accuracy and reducing bias in AI algorithms. It should be stressed that the greater access to data that gives FL an advantage over centralized learning rather than any inherent algorithmic benefit. Sheller \textit{et al.} previously showed that FL can achieve similar accuracy as centralized learning but may be superior to similar collaborative learning techniques and to training on data from a single institution \cite{sheller2019, sheller2020}.

\section{Conclusion and future work}

We have introduced Open Federated Learning (OpenFL)\footnote{\url{https://github.com/intel/openfl}}, as a production-ready FL package that allows developers to train ML models on the nodes of remote data owners. The OpenFL interface makes it easy for data scientists to port their existing ML models, whether in TensorFlow, PyTorch, or some other ML library, into a distributed training pipeline. While it was created to address problems discovered in academia, it has now being adopted by companies because of its unique security focus. The development of OpenFL has benefited significantly from our external collaborations, and by making the project open source we hope that it will continue to be shaped by the wider FL community in new and exciting ways. Our goal with OpenFL is not to compete with other FL open-source software efforts, but to inter-operate and collaborate towards providing a comprehensive solution for data-private collaborative learning.

Our ambition is that federations, such as the FeTS Initiative\footnote{\url{https://www.fets.ai}}, will not serve as \textit{ad hoc} collaborations for specific research efforts, but will serve as permanent networks for researchers in the healthcare, financial, industrial, and retail industries to more effectively train, deploy, monitor, and update their AI algorithms over time.

\section*{Acknowledgments}
Research reported in this publication was partly supported by the National Institutes of Health (NIH) under award number NIH/NCI:U01CA242871. The content of this publication is solely the responsibility of the authors and does not represent the official views of the NIH.

\bibliographystyle{ieeetr}
\bibliography{bibliography.bib}

\begin{thebibliography}{10}

\bibitem{paullada2020data}
A.~Paullada, I.~D. Raji, E.~M. Bender, E.~Denton, and A.~Hanna, ``Data and its
  (dis) contents: A survey of dataset development and use in machine learning
  research,'' {\em arXiv preprint arXiv:2012.05345}, 2020.

\bibitem{sheller2020}
M.~J. Sheller, B.~Edwards, G.~A. Reina, J.~Martin, S.~Pati, A.~Kotrotsou,
  M.~Milchenko, W.~Xu, D.~Marcus, R.~R. Colen, and S.~Bakas, ``Federated
  learning in medicine: facilitating multi-institutional collaborations without
  sharing patient data,'' {\em Sci Rep.}, vol.~10, no.~1, p.~12598, 2020.

\bibitem{sheller2019}
M.~J. Sheller, G.~A. Reina, B.~Edwards, J.~Martin, and S.~Bakas,
  ``Multi-institutional deep learning modeling without sharing patient data: A
  feasibility study on brain tumor segmentation,'' {\em Brainlesion},
  vol.~11383, pp.~92--104, 2019.

\bibitem{yang2019federated}
Q.~Yang, Y.~Liu, T.~Chen, and Y.~Tong, ``Federated machine learning: Concept
  and applications,'' {\em ACM Trans. Intell. Syst. Technol.}, vol.~10, Jan.
  2019.

\bibitem{mcmahan2017communicationefficient}
B.~McMahan, E.~Moore, D.~Ramage, S.~Hampson, and B.~A. y~Arcas,
  ``Communication-efficient learning of deep networks from decentralized
  data,'' in {\em Artificial Intelligence and Statistics}, pp.~1273--1282,
  PMLR, 2017.

\bibitem{annas2003hipaa}
G.~J. Annas {\em et~al.}, ``Hipaa regulations-a new era of medical-record
  privacy?,'' {\em New England Journal of Medicine}, vol.~348, no.~15,
  pp.~1486--1490, 2003.

\bibitem{voigt2017eu}
P.~Voigt and A.~Von~dem Bussche, ``The eu general data protection regulation
  (gdpr),'' {\em A Practical Guide, 1st Ed., Cham: Springer International
  Publishing}, vol.~10, p.~3152676, 2017.

\bibitem{suzumura2019federated}
T.~Suzumura, Y.~Zhou, N.~Baracaldo, G.~Ye, K.~Houck, R.~Kawahara, A.~Anwar,
  L.~L. Stavarache, Y.~Watanabe, P.~Loyola, {\em et~al.}, ``Towards federated
  graph learning for collaborative financial crimes detection,'' {\em arXiv
  preprint arXiv:1909.12946}, 2019.

\bibitem{tensorflow}
M.~Abadi, P.~Barham, J.~Chen, Z.~Chen, A.~Davis, J.~Dean, M.~Devin,
  S.~Ghemawat, G.~Irving, M.~Isard, {\em et~al.}, ``Tensorflow: A system for
  large-scale machine learning,'' in {\em 12th $\{$USENIX$\}$ symposium on
  operating systems design and implementation ($\{$OSDI$\}$ 16)}, pp.~265--283,
  2016.

\bibitem{pytorch}
A.~Paszke, S.~Gross, F.~Massa, A.~Lerer, J.~Bradbury, G.~Chanan, T.~Killeen,
  Z.~Lin, N.~Gimelshein, L.~Antiga, {\em et~al.}, ``Pytorch: An imperative
  style, high-performance deep learning library,'' in {\em Advances in neural
  information processing systems}, pp.~8026--8037, 2019.

\bibitem{bonawitz2019federated}
K.~Bonawitz, H.~Eichner, W.~Grieskamp, D.~Huba, A.~Ingerman, V.~Ivanov,
  C.~Kiddon, J.~Kone{\v{c}}n{\`y}, S.~Mazzocchi, H.~B. McMahan, {\em et~al.},
  ``Towards federated learning at scale: System design,'' {\em arXiv preprint
  arXiv:1902.01046}, 2019.

\bibitem{grpc}
X.~Wang, H.~Zhao, and J.~Zhu, ``Grpc: A communication cooperation mechanism in
  distributed systems,'' {\em ACM SIGOPS Operating Systems Review}, vol.~27,
  no.~3, pp.~75--86, 1993.

\bibitem{kairouz2021advances}
P.~Kairouz, H.~B. McMahan, B.~Avent, A.~Bellet, M.~Bennis, A.~N. Bhagoji,
  K.~Bonawitz, Z.~Charles, G.~Cormode, R.~Cummings, {\em et~al.}, ``Advances
  and open problems in federated learning,'' {\em arXiv preprint
  arXiv:1912.04977}, 2019.

\bibitem{gsgx}
C.~che Tsai, D.~E. Porter, and M.~Vij, ``Graphene-sgx: A practical library {OS}
  for unmodified applications on {SGX},'' in {\em 2017 {USENIX} Annual
  Technical Conference ({USENIX} {ATC} 17)}, (Santa Clara, CA), pp.~645--658,
  {USENIX} Association, July 2017.

\bibitem{mnist}
S.~Zhang, ``Mnist data,'' 2020.

\bibitem{pati2021gandlf}
S.~Pati, S.~P. Thakur, M.~Bhalerao, U.~Baid, C.~Grenko, B.~Edwards, M.~Sheller,
  J.~Agraz, B.~Baheti, V.~Bashyam, {\em et~al.}, ``Gandlf: A generally nuanced
  deep learning framework for scalable end-to-end clinical workflows in medical
  imaging,'' {\em arXiv preprint arXiv:2103.01006}, 2021.

\bibitem{labelfusion}
S.~Pati, ``{LabelFusion: Medical Image label fusion of segmentations},'' Mar.
  2021.

\bibitem{brats_1}
B.~H. Menze, A.~Jakab, S.~Bauer, J.~Kalpathy-Cramer, K.~Farahani, J.~Kirby,
  Y.~Burren, N.~Porz, J.~Slotboom, R.~Wiest, {\em et~al.}, ``The multimodal
  brain tumor image segmentation benchmark (brats),'' {\em IEEE transactions on
  medical imaging}, vol.~34, no.~10, pp.~1993--2024, 2014.

\bibitem{brats_2}
S.~Bakas, H.~Akbari, A.~Sotiras, M.~Bilello, M.~Rozycki, J.~S. Kirby, J.~B.
  Freymann, K.~Farahani, and C.~Davatzikos, ``Advancing the cancer genome atlas
  glioma mri collections with expert segmentation labels and radiomic
  features,'' {\em Scientific data}, vol.~4, no.~1, pp.~1--13, 2017.

\bibitem{brats_3}
S.~Bakas, M.~Reyes, A.~Jakab, S.~Bauer, M.~Rempfler, A.~Crimi, R.~T. Shinohara,
  C.~Berger, S.~M. Ha, M.~Rozycki, {\em et~al.}, ``Identifying the best machine
  learning algorithms for brain tumor segmentation, progression assessment, and
  overall survival prediction in the brats challenge,'' {\em arXiv preprint
  arXiv:1811.02629}, 2018.

\bibitem{langlotz_bias}
A.~Kaushal, R.~Altman, and C.~Langlotz, ``Health care ai systems are biased,''
  {\em Scientific American (Online)}, November 17, 2020.
\newblock
  \url{https://www.scientificamerican.com/article/health-care-ai-systems-are-biased/}.

\bibitem{gender_imbalance_ai}
A.~J. Larrazabal, N.~Nieto, V.~Peterson, D.~H. Milone, and E.~Ferrante,
  ``Gender imbalance in medical imaging datasets produces biased classifiers
  for computer-aided diagnosis,'' {\em Proceedings of the National Academy of
  Sciences}, vol.~117, no.~23, pp.~12592--12594, 2020.

\bibitem{pmlr-v81-buolamwini18a}
J.~Buolamwini and T.~Gebru, ``Gender shades: Intersectional accuracy
  disparities in commercial gender classification,'' in {\em Proceedings of the
  1st Conference on Fairness, Accountability and Transparency} (S.~A. Friedler
  and C.~Wilson, eds.), vol.~81 of {\em Proceedings of Machine Learning
  Research}, (New York, NY, USA), pp.~77--91, PMLR, 23--24 Feb 2018.

\bibitem{coston_covariate}
A.~Coston, K.~N. Ramamurthy, D.~Wei, K.~R. Varshney, S.~Speakman, Z.~Mustahsan,
  and S.~Chakraborty, ``Fair transfer learning with missing protected
  attributes,'' in {\em Proceedings of the 2019 AAAI/ACM Conference on AI,
  Ethics, and Society}, AIES '19, (New York, NY, USA), p.~91–98, Association
  for Computing Machinery, 2019.

\bibitem{rieke2020}
N.~Rieke, J.~Hancox, W.~Li, F.~Milletarì, H.~R. Roth, S.~Albarqouni, S.~Bakas,
  M.~N. Galtier, A.~L. Bennett, K.~Maier-Hein, S.~B. Ourselin, M.~Sheller,
  R.~M. Summers, A.~Trask, D.~Xu, Maximilian, and M.~J. Cardoso, ``The future
  of digital health with federated learning,'' {\em npj Digit. Med.}, vol.~3,
  no.~119, 2020.

\end{thebibliography}

\end{document}